\documentclass{article}

\usepackage[preprint,nonatbib]{neurips_2020}

\usepackage[utf8]{inputenc} 
\usepackage[T1]{fontenc}    
\usepackage{hyperref}       
\usepackage{url}            
\usepackage{booktabs}       
\usepackage{amsfonts}       
\usepackage{nicefrac}       
\usepackage{microtype}      

\usepackage{amsmath}
\usepackage{tabularx}

\DeclareMathOperator*{\argmax}{arg\,max}

\newcommand{\vecX}{\mathbf{x}}
\newcommand{\vecY}{\mathbf{y}}
\newcommand{\vecZ}{\mathbf{z}}
\newcommand{\vecH}{\mathbf{h}}
\newcommand{\vecP}{\mathbf{p}}
\newcommand{\vecT}{\boldsymbol{\theta}}
\newcommand{\matP}{\mathbf{P}}

\usepackage[]{algorithm2e}
\usepackage{graphicx}
\usepackage{xcolor, colortbl}
\definecolor{Gray}{gray}{0.9}

\usepackage[numbers]{natbib}
\title{Differentiable Language Model Adversarial Attacks on Categorical Sequence Classifiers}

\author{%
  I.~Fursov \\
  \texttt{ivan.fursov@skoltech.ru} \\
   \And
   A.~Zaytsev \\
   \texttt{a.zaytsev@skoltech.ru} \\
   \And
   N.~Kluchnikov \\
   \texttt{nikita.klyuchnikov@skoltech.ru} \\
   \And
   A.~Kravchenko \\
   \texttt{andrey.kravchenko@deepreason.ai} \\
   \And
   E.~Burnaev \\
   \texttt{e.burnaev@skoltech.ru} \\
}

\newenvironment{itemizetight}
{\vspace{-.3\baselineskip}\begin{itemize}\setlength{\itemsep}{-.15\baselineskip}}
{\end{itemize}\vspace{-.3\baselineskip}}

\begin{document}

\maketitle

\begin{abstract}
An adversarial attack paradigm explores various scenarios for the vulnerability of deep learning models: minor changes of the input can force a model failure. Most of the state of the art frameworks focus on adversarial attacks for images and other structured model inputs, but not for categorical sequences models. 

Successful attacks on classifiers of categorical sequences are challenging because the model input is tokens from finite sets, so a classifier score is non-differentiable with respect to inputs, and gradient-based attacks are not applicable. 
Common approaches deal with this problem working at a token level, while the discrete optimization problem at hand requires a lot of resources to solve.

We instead use a fine-tuning of a language model for adversarial attacks as a generator of adversarial examples. 
To optimize the model, we define a differentiable loss function that depends on a surrogate classifier score and on a deep learning model that evaluates approximate edit distance. So, we control both the adversability of a generated sequence and its similarity to the initial sequence.

As a result, we obtain semantically better samples. Moreover, they are resistant to adversarial training and adversarial detectors. Our model works for diverse datasets on bank transactions, electronic health records, and NLP datasets.
\end{abstract}

\section{Introduction}
\label{sec:intro}

Adversarial attacks~\cite{yuan2019adversarial} in all application areas including computer vision~\cite{akhtar2018threat,khrulkov2018art}, NLP~\cite{zhang2019adversarial,wang2019survey}, and graphs~\cite{sun2018adversarial} make use of non-robustness of deep learning models.
An adversarial attack generates an object that fools a deep learning model by inserting changes into the initial object, undetectable by a human eye. The deep learning model misclassifies the generated object, whilst for a human, it is evident that the object's class remains the same~\cite{Kurakin2017}.

For images, we can calculate derivatives of class probabilities with respect to the pixels' colors in an input image.
Thus, moving along the gradient we apply slight alterations to a few pixels and get a misclassified adversarial image close to the input image.
If we have access to the gradient information, it is easy to construct an adversarial example.

The situation is different for discrete sequential data, due to its discrete categorical nature~\cite{zhang2019adversarial,wang2019survey}.
Whilst an object representation can lie in a continuous space, when we go back to the space of sequences, we can move far away from the initial object, rendering partial derivatives useless. 
Many approaches that accept the initial space of tokens as input attempt to modify these sequences using operations like addition, replacement, or switching of tokens~\cite{samanta2017towards, liang2017deep,ebrahimi2018hotflip}. 
The discrete optimisation problem is often hard, and existing approaches are greedy or similar ones~\cite{ebrahimi2018hotflip}.
Another idea is to move into an embedding space and leverage on gradients and optimisation approaches in that space~\cite{sato2018interpretable,ren2020generating}.
This approach requires training of a large model from scratch and has many parameters to tune.
Moreover, in this type of models without attention mechanism the embedding space is a bottleneck, as we constrain all information about input sequence to the embedding vector.
Thus, this type of models is inaccurate for long and complex sequences.

\begin{figure}[!ht]
    \centering
  \includegraphics[width=0.5\textwidth]{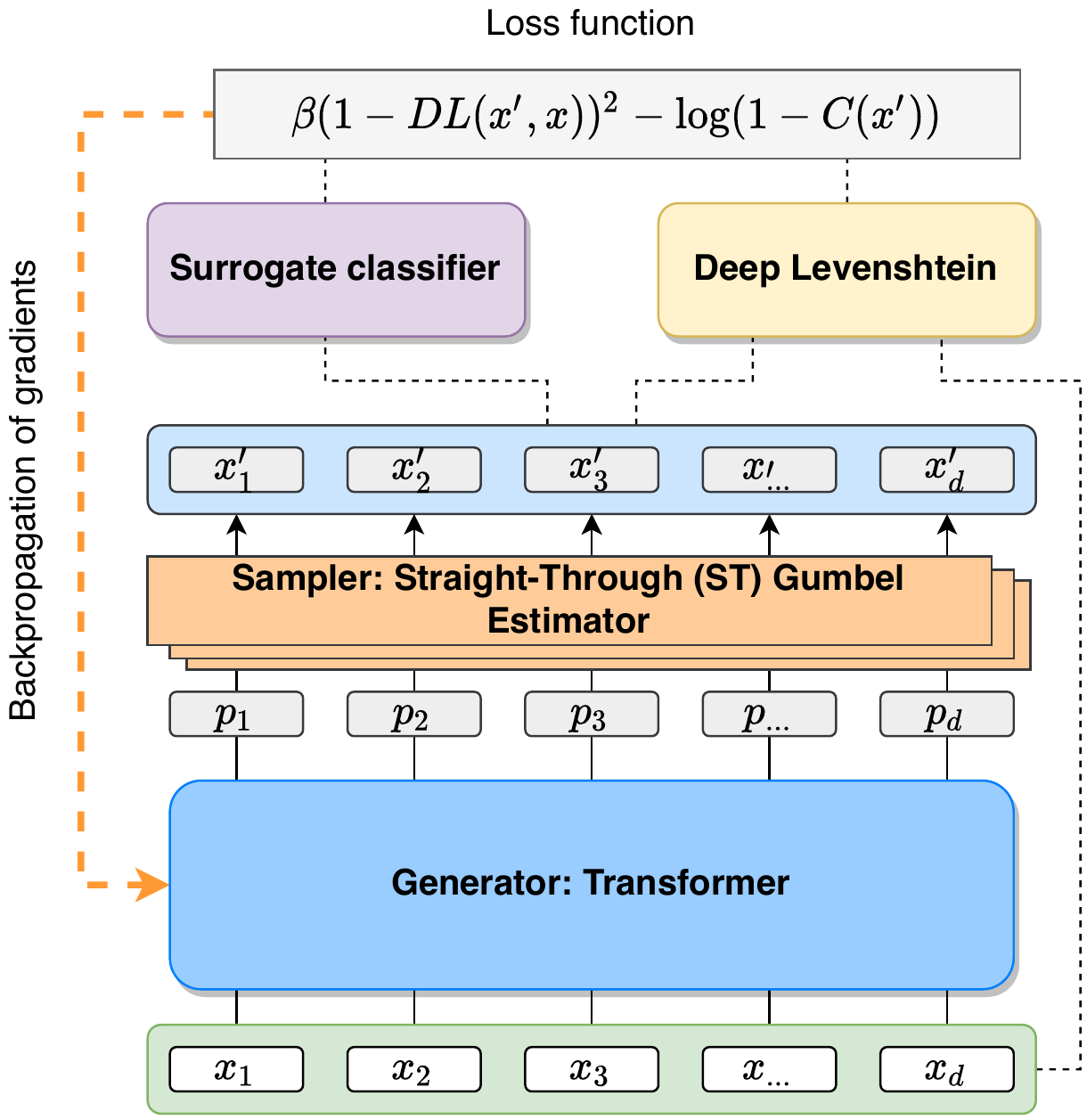}
  \caption{
  Training of the DILMA architecture consists of the following steps. \textbf{Step 1}: obtain logits $\matP$ from a pre-trained Language Model for input $\vecX$. \textbf{Step 2}: sample $\vecX'$ from multinomial distribution $\matP$ using the Gumbel-Softmax estimator. To improve generation quality we can sample many times. \textbf{Step 3}: obtain surrogate probability $C(\vecX')$ and approximate edit distance $DL(\vecX', \vecX)$. \textbf{Step 4}: calculate loss, do a backward pass. \textbf{Step 5}: update parameters of the language model using these gradients. }
    \label{fig:architecture}
\end{figure}

We propose an adversarial attack model DILMA (DIfferentiable Language Model Attack) that can alleviate the aforementioned problems with differentiability.
Moreover, our approach benefits from strong pre-trained language models, making adversarial model generation easier for new domains.
Our model for an adversarial attack works in two regimes. The first regime is a random sampling that produces adversarial examples by chance. The second regime is a targeted attack that modifies the language model by optimisation of the loss related to misclassification by the target model and small difference between an initial sequence and its adversarial counterpart.
For the second regime we introduce a differentiable loss function as the weighted sum of the distance between the initial sequence and the generated one and the difference between the probability scores for these sequences.
We use a trained differentiable version of the Levenshtein distance~\cite{moon2018multimodal} and Gumbel-Softmax heuristic to pass the derivatives through our sequence generation layers.
The number of hyperparameters in our method is small, and the selection of them is not crucial for obtaining high performance scores.
As our loss is differentiable, we can adopt any gradient-based adversarial attack.
The training and inference procedure is summarised in Figure~\ref{fig:architecture}.

As a generative model for adversarial attacks we use a transformer sequence2sequence masked language model based on BERT~\cite{devlin2018bert}, thus allowing the constructed model to be reused for generating adversarial attacks.
The validation of our approaches includes testing on diverse datasets from NLP, bank transactions, and electronic healthcare records domains.
Examples of generated sequences for the TREC dataset \cite{Voorhees1999TheTQ} are presented in Table~\ref{tab:adversarial_examples}.

\begin{table}[]
\begin{tabularx}{\textwidth}{lll}
\toprule
\textbf{Original}                    & \textbf{Hotflip}                      & \textbf{DILMA (ours)}                     \\ \midrule
how did socrates die                                                              & \textcolor{purple}{fear} did socrates die                                                             & how did \textcolor{purple}{jaco} die                                                                \\ 
\rowcolor[HTML]{D9D2D1}
what were the 
& what were the
& what were the \\ 
\rowcolor[HTML]{D9D2D1}
first frozen foods 
& \textcolor{purple}{origin} frozen foods                                 
& \textcolor{purple}{second} frozen foods \\ 
what country's people are                            & what country \textcolor{purple}{why} caused are                           & what \textcolor{purple}{company} s people are                            \\ 
the top television watchers                             
&  the top television watchers
& the top television watchers \\ 
\rowcolor[HTML]{D9D2D1}
what is the difference between
& what \textcolor{purple}{fear} the which between
& what is the \textcolor{purple}{novel} between \\
\rowcolor[HTML]{D9D2D1}
a median and a mean 
& a median and a mean 
& a \textcolor{purple}{bachelor and a play} \\
what s the literary term 
& what s the \textcolor{purple}{origin} term
& what s the \textcolor{purple}{human} term   \\ 
for a play on words
& for a play on words
& for a play on words  \\
\bottomrule                                     

\end{tabularx}
\caption{Examples of generated adversarial sequences for the TREC dataset \cite{Voorhees1999TheTQ} evaluated on the HotFlip and DILMA approaches. HotFlip often selects \textbf{origin} and \textbf{fear} words that corrupt the semantics of a sentence. DILMA is more ingenious.}
\label{tab:adversarial_examples}
\end{table}

To sum up, the main contributions of this work are the following:
\begin{itemizetight}
    \item An adversarial attack based on a masked language model (MLM). We fine tune parameters of MLM by optimising a weighted sum of two differentiable terms based on a surrogate distance between sequences and a surrogate classifier model scores.
    \item A simple, but powerful baseline ``SamplingFool'' attack that requires only a pre-trained MLM to work.
    \item Our algorithms are resistant to common defense strategies, whilst existing approaches fail to fool models after they defende themselves. To validate our approach we use diverse datasets from NLP, bank transactions, and electronic health records areas.
    \item We show a necessity of validation adversarial attacks for sequential data using different defense strategies. Some state-of-the-art approaches lack the ability to overcome simple defences. Our methods perform well in these scenarios.
\end{itemizetight}

\section{Related work}

There exist adversarial attacks for different types of data:
the most mainstream being images data~\cite{szegedy2014intriguing,goodfellow2014explaining}, graph data,~\cite{zugner2018adversarial} and sequences~\cite{papernot2016crafting}.
We refer a reader to recent surveys on adversarial attacks for sequences~\cite{zhang2019adversarial,wang2019survey}, and highlight below only related issues.

\cite{papernot2016crafting} is one of the first published works on generation of adversarial attacks for discrete sequences.
The authors identify two main challenges for adversarial attacks on discrete sequence models: a discrete space of possible objects and a complex definition of a semantically coherent sequence.
Their approach focuses on a white-box adversarial attack for binary classification.
We consider a similar problem statement,
but focus on black-box adversarial attacks for sequences, see also~\cite{gao2018black,liang2017deep,jin2020is}. 
Authors of~\cite{gao2018black} identify certain pairs of tokens and then permute their positions within those pairs, thus working directly on a token level.
A black-box approach~\cite{liang2017deep} also performs a direct search for the most harmful tokens to fool a classifier.

As stated in Section~\ref{sec:intro} there are two main classes of approaches. In the first one they directly apply natural modifications like deletion or replacement of a token in a sequence trying to generate a good adversarial example~\cite{samanta2017towards,xu2020texttricker}.
This idea leads to an extensive search among the space of possible sequences, thus making the problem computationally challenging, especially if the inference time of a model is significant~\cite{fursov2019sequence}.
Moreover, we have little control on the semantic closeness of the initial and modified sequences. An approach~\cite{cheng2018seq2sick} to white-box adversarial attacks is very similar except one more modification: the authors additionally use differentiable distance between the initial and generated sequences to control their similarity.



The second type of methods is similar to the common approaches used by adversarial attacks practitioners, namely, to use gradients of a cost function to modify the initial sequence.
However, the authors in~\cite{sato2018interpretable} limit directions of perturbations in an embedding space by modifying only a specific token,  which seems too restrictive. 
\cite{ren2020generating} elaborates a VAE-based model with GRU units inside to create a generator for adversarial examples. However, due to usage of RNNs seq2seq models complexity and length of sequences they can process is limited.
In our opinion, the state of the art mechanisms such as attention~\cite{vaswani2017attention} should be used instead of standard RNNs to improve seq2seq models. 
Authors of~\cite{fursov2020gradient} move in the right direction, but the performance metrics are worse than of competitors'.





From the current state of the art, we see a remaining need to identify an effective way to generate adversarial categorical sequences.
Existing approaches use previous generations of LMs based on recurrent architectures, and stay at a token level or use VAE, despite known limitations of these models~\cite{yang2017improved,vaswani2017attention}. Moreover, as most of the applications focus on NLP-related tasks, it makes sense to widen the scope of application domains for adversarial attacks on categorical sequences. 


\section{Methods}

\subsection{General description of the approach} 
We generate adversarial examples using two consecutive components: a masked language model (MLM, we call it language model or LM below for brevity) with parameters $\vecT$ that provides for an input sequence $\vecX$, conditional distribution $p_{\vecT}(\vecX'|\vecX)$, and a sampler from this distribution such that $\vecX' \sim p_{\vecT}(\vecX'|\vecX)$.
Thus, we can generate sequences $\vecX'$ by consecutive application of a LM and a sampler.

For this sequence to be adversarial, we optimise a proposed differentiable loss function that forces the LM to generate semantically similar but adversarial examples by modifying the LM parameters $\vecT$.
The loss function consists of two terms: the first term corresponds to a surrogate classifier $C(\vecX')$ that outputs a probability of belonging to a target class, and the second term corresponds to a Deep Levenstein distance $DL(\vecX, \vecX')$ that approximates the edit distance between sequences.

The general scheme of our approach is given in Figure~\ref{fig:architecture}.
More details are given below.
We start with the description of the LM and the sampler in Subsection~\ref{sec:seq2seqmodel}.
We continue with the description of the loss function in Subsection~\ref{sec:loss_function}.
The formal description of our algorithm is given in Subsection~\ref{sec:dilma_description}.
In later subsections~\ref{sec:classifiers} and~\ref{sec:deep_levenstein} we provide more details on the use of the target and surrogate classifiers and the Deep Levenstein model correspondingly.
Detailed descriptions of the architectures and training procedures is provided in supplementary materials~\ref{sec:classifiers_sup}.

\subsection{Language model} 
\label{sec:seq2seqmodel}

The language model (LM) lies at the heart of our approach.
The LM is a model that takes a sequence of tokens (e.g. words) as input $\vecX = \{x_1, \ldots, x_t\}$ and outputs logits for tokens $\matP = \{\vecP_1, \ldots, \vecP_{t}\} \in \mathbb{R}^{d \cdot t}$ for each index $1, \ldots, t$, where $d$ is a size of the dictionary of tokens. In this work we use transformer architecture as the LM~\cite{vaswani2017attention}.
We pre-train a transformer encoder \cite{vaswani2017attention} in a BERT \cite{devlin2018bert} manner on a corresponding domain. We use all available data to train such this kind of model.

The sampler is defined as follows: the LM learns a probability distribution over sequences, so we can sample a new sequence $\vecX' = \{x'_1, \ldots, x'_t\}$ based on the vectors of token logits $\matP$. In this work we use Straight-Through Gumbel Estimator $ST(\matP): \matP \rightarrow \vecX'$ for sampling~\cite{jang2017categorical}.
To get the actual probabilities $q_{ij}$ from logits $p_{ij}$
we use a softmax with temperature $\tau$ \cite{hinton2015distilling}:
    \begin{equation}
    \label{eqeq}
        q_{ij} = \frac{\exp \left(p_{ij} / \tau \right)}{\sum_{k = 1}^d \exp \left(p_{ik} / \tau \right)}.
    \end{equation}
Value $\tau > 1$ produces a softer probability distribution over classes. As $\tau \rightarrow \infty$, the original distribution approaches a uniform distribution. If $\tau\to0$, then sampling from \eqref{eqeq} reduces to setting $x'_i = \argmax_j \vecP_i = \argmax \{p_{i1}, \ldots, p_{id}\}^{\top}$.

Our first method \textbf{SamplingFool} samples sequences from the categorical distribution with $q_{ij}$ probabilities. The sampled examples turn out to be good-looking and can easily fool a classifier, as we discuss in Section \ref{sec:experiments}. The method is similar to the random search algorithm and serves as a baseline for the generation of adversarial examples for discrete sequences.

\subsection{Loss function}
\label{sec:loss_function}

The Straight-Through Gumbel sampling allows propagation of the derivatives through the softmax layer.
So, we optimise parameters $\vecT$ of our MLM to improve quality of generated adversarial examples.

A loss function should take into account two terms: the first term represents the probability score drop $(1 - C^t_y(\vecX'))$ of the target classifier for the considered class $y$, the second term represents the edit distance $DL(\vecX, \vecX')$ between the initial sequence $\vecX$ and the generated sequence $\vecX'$. 
We should maximise the probability drop and minimise the edit distance, so it is as close to $1$ as possible.

In our black-box scenario we do not have access to the true classifier score, so we use a substitute classifier score $C_y(\vecX') \approx C^t_y(\vecX')$.
More details on classifiers are given in Subsection~\ref{sec:classifiers}.
As a differentiable alternative to edit distance, we use Deep Levenstein model proposed in \cite{moon2018multimodal} --- a deep learning model that approximates the edit distance: $DL(\vecX, \vecX') \approx D(\vecX, \vecX')$.
More details on the Deep Levenstein model are given in Subsection~\ref{sec:deep_levenstein}.

In that way we get the following differentiable loss function:
\begin{equation}\label{eq:dilma_loss}
    L(\vecX', \vecX, y) = \beta (1 - DL(\vecX', \vecX))^2 - \log (1 - C_y(\vecX')), 
\end{equation}
where $C_y(\vecX')$ is the probability of the true class $y$ for sequence $\vecX'$ and $\beta$ is a weighting coefficient. Thus, we penalise cases when more than one modification is needed in order to get $\vecX'$ from $\vecX$. Since we focus on non-target attacks, the $C_y(\vecX')$ component is included in the loss. The smaller the probability of an attacked class, the smaller the loss.



We estimate derivatives of $L(\vecX', \vecX, y)$ in a way similar to~\cite{jang2016categorical}.
Using these derivatives we do a backward pass and update the weights $\vecT$ of the Language Model. We find that updating the whole set of parameters $\vecT$ is not the best strategy and a better alternative is to update only the last linear layer and the last layer of the transformer.

\subsection{DILMA algorithm}
\label{sec:dilma_description}

Now we are ready to group the introduced components into a formal algorithm with the architecture depicted in Figure~\ref{fig:architecture}.

The proposed approach for the generation of adversarial sequences has the following inputs:
a language model with parameters $\vecT = \vecT_0$, learning rate $\alpha$, temperature for sampling $\tau$, coefficient $\beta$ (if $\beta = 0$, we only decrease a classifier score and don't pay attention to the edit distance part of the loss function) for the normalised classifier score in~\eqref{eq:dilma_loss}.
For these inputs the algorithm performs the following steps at iterations $i = 1, 2, \ldots, k$ for a given sequence $\vecX$.
\begin{itemizetight}
    \item[\textbf{Step 1.}] Pass the sequence $\vecX$ through the pre-trained LM. Obtain logits $\matP = LM_{\vecT_{i - 1}}(\vecX)$.
    \item[\textbf{Step 2.}] Sample an adversarial sequence $\vecX'$ from the logits using the Gumbel-Softmax Estimator.
    \item[\textbf{Step 3.}] Calculate the probability $C_y(\vecX')$ and Deep Levenstein distance $DL(\vecX', \vecX)$. Calculate the loss value $L(\vecX', \vecX, y)$~\eqref{eq:dilma_loss}.
    \item[\textbf{Step 4.}] Do a backward pass to update LM's weights $\vecT_{i-1}$ using gradient descent and get new weights $\vecT_{i}$.
    \item[\textbf{Step 5.}] Obtain an adversarial sequence $\vecX'_i$.
    \begin{itemizetight}
        \item[a)] \textbf{DILMA}: by setting $\vecX'_i$ to be the most probable value according to $\matP  = LM_{\vecT_{i}}(\vecX)$, or
        \item[b)] \textbf{DILMA w/ sampling}: by sampling based on \eqref{eqeq} with $\tau>0$ and $\matP$.
    \end{itemizetight}
\end{itemizetight}

Note that the algorithm decides by itself which tokens should be replaced.  The classification component changes the sequence in a direction where the probability score $C_y(\vecX')$ is low, and the Deep Levenstein distance keeps the generated sequence close to the original one.
The update procedure for each $\vecX$ starts from the pre-trained LM parameters $\vecT_0$. 

Let us note that for the \textbf{DILMA w/ sampling} approach we can sample $m>1$  adversarial examples for each iteration with almost no additional computational cost. As we will see in Section \ref{sec:experiments}, this approach works better than the original \textbf{DILMA} approach across all datasets.


After all iterations $i = 1, 2, \ldots, k$ of the algorithm we obtain a set of adversarial sequences $\{\vecX_i'\}_{i = 1}^k$ in case of the original \textbf{DILMA}; for \textbf{DILMA w/ sampling} we get $m$ adversarial sequences on each iteration of the algorithm. The last sequence in this set is not always the best one. Therefore among generated sequences that are adversarial w.r.t. the substitute classifier $C_y(\vecX')$ we select $\vecX'_{\mathrm{opt}}$ with the lowest  Word Error Rate (WER). Here we calculate WER relative to the initial sequence $\vecX$.
If all examples are not adversarial w.r.t. $C_y(\vecX')$, then we select $\vecX'_{\mathrm{opt}}$ with the smallest target class score. 
As we use only the substitute classifier, \textbf{DILMA} is a blackbox-type adversarial attack.


\subsection{Classification model}
\label{sec:classifiers}

In all experiments we use two classifiers: a target classifier $C^t(\vecX)$ that we attack and a substitute classifier $C(\vecX)$ that provides differentiable surrogate classifier scores. 

The target classifier is a combination of a bi-directional Gated Recurrent Unit (GRU)~\cite{chung2014empirical} RNN and an embeddings layer for tokens before it. 
The hidden size for GRU is $128$, the dropout rate is $0.1$ and the embedding size $100$.
The surrogate classifier is Convolutional Neural Networks (CNN) for Sentence Classification~\cite{kim2014convolutional} and an embeddings layer of size $100$ for tokens before it.

The substitute classifier has access only to $50\%$ of the data, whilst the target classifier uses the whole dataset. We split the dataset into two parts with stratification. Both models demonstrate comparable results for all datasets.

\subsection{Deep Levenstein}
\label{sec:deep_levenstein}

To make gradient-based updates of parameters, we need a differentiable version of the edit distance function. 
We use an approach based on training a deep learning model ``Deep Levenshtein distance'' $DL(\vecX, \vecX')$ for evaluation of the Word Error Rate between two sequences $\vecX$ and $\vecX'$. 
It is similar to the approach proposed in \cite{moon2018multimodal}. 
In our case, the WER is used instead of the Levenstein distance, since we work at the word level instead of the character level for NLP tasks, and for non-textual tasks there are no levels other than ``tokens''.

The Deep Levenstein model receives two sequences $(\vecX, \vecY)$. 
It encodes them into a dense representation of fixed length $l$ using the shared encoder $\vecZ_\vecX = E(\vecX)$, $\vecZ_\vecY = E(\vecY)$. Then it concatenates the representations and the absolute difference between them in a vector $(\vecZ_\vecY, \vecZ_\vecY, |\vecZ_\vecY - \vecZ_\vecY|)$ of length $3l$. 
At the end the model uses a fully-connected layer.
The architecture is similar to the one proposed in~\cite{dai2020convolutional}. 
To estimate the parameters of the encoder and the fully connected layer we use the $L_2$ loss between true and model values of the WER distance.
We form a training sample of size of about two million data points by sampling sequences and their modifications from the training data.

\begin{table}[]
    \centering
    \begin{tabular}{cccccccc}
    \hline
& Classes & Avg.  & Max  & Train  & Test & Targeted GRU & Substitute CNN \\
& & length & length & size & size & accuracy & accuracy \\
\hline
AG & 4 & 6.61 & 19 & 120000 & 7600 & 0.87 & 0.86 \\
TREC & 6 & 8.82 & 33 & 5452 & 500 & 0.85 & 0.85 \\
SST-2 & 2 & 8.62 & 48 & 76961 & 1821 & 0.82 & 0.80 \\
MR & 2 & 18.44 & 51 & 9595 & 1067 & 0.76 & 0.70 \\
\hline
EHR & 2 & 5.75 & 20 & 314724 & 34970 & 0.99 & 0.98 \\
Tr.Age & 4 & 8.98 & 42 & 2649496 & 294389 & 0.46 & 0.45 \\
Tr.Gender & 2 & 10.26 & 20 & 256325 & 28481 & 0.68 & 0.67 \\
\hline
    \end{tabular}
    \caption{The comparison of evaluated datasets. We try to attack classifiers for diverse NLP- and non-NLP problems}
    \label{table:datasets_comparison}
\end{table}

\section{Experiments}
\label{sec:experiments}

In this section we describe our experiments. The datasets and the source code are published online\footnote{The code is available at \url{https://github.com/fursovia/dilma}}.

An ablation study and an algorithm to select hyperparameters as well as additional experiments are provided in supplementary materials in Section~\ref{sec:additional_experiments}.

\subsection{Competitor approaches}

We have compared our approach to HotFlip, FGSM variant for adversarial sequences~\cite{goodfellow2014explaining, papernot2016crafting}, and DeepFool~\cite{moosavidezfooli2015deepfool}.
We have also tried to implement~\cite{ren2020generating}, but haven't managed to find hyperparameters that provide performance similar to that reported by the authors. 

\textbf{HotFlip}. The main idea of HotFlip~\cite{ebrahimi2018hotflip} is to select the best token to change, given an approximation of partial derivatives for all tokens and all elements of the dictionary.
We change multiple tokens in a greedy way or by a beam search of a good sequence of changes.

\textbf{FGSM} chooses random token in a sequence and uses the Fast Gradient Sign Method to perturb its embedding vector. Then, the algorithm finds the closest vector in an embedding matrix and replaces the chosen token with the one that corresponds to the identified vector.

\textbf{DeepFool} follows the same idea of gradient-based replacement method in an embedded space.
In addition, by assuming the local linearity of a classifier DeepFool provides a heuristic to select the most promising modification.

\subsection{Datasets}

We have conducted experiments on four open NLP datasets for text classification and three non-NLP datasets (bank transactions and electronic health records datasets). The datasets’ statistics are listed in Table~\ref{table:datasets_comparison}.

\textbf{The AG News corpus} (AG) \cite{zhang2015characterlevel} consists of news articles on the web from the AG corpus. It has four classes: World, Sports, Business, and Sci/Tech. Both training and test sets are perfectly balanced.
\textbf{The TREC dataset} (TREC) \cite{Voorhees1999TheTQ} is a dataset for text classification consisting of open-domain, fact-based questions divided into broad semantic categories. We use a six-class version (TREC-6). 
\textbf{The Stanford Sentiment Treebank} (SST-2) \cite{socher-etal-2013-recursive} contains phrases with fine-grained sentiment labels in the parse trees of $11,855$ sentences from movie reviews.
\textbf{The Movie Review Data} (MR) \cite{Pang_43} is a movie-review data set of sentences with sentiment labels (positive or negative).

\textbf{Electronic Health Records} (EHR) \cite{fursov2019sequence} dataset contains information on medical treatments. The goal is to help an insurance company to detect frauds based on a history of visits of patients to a doctor. Each sequence consists of visits with information on a drug code and the amount of money spent at each visit.
This dataset is imbalanced, as the share of frauds is $1.5\%$.

We have also used two open bank \textbf{Transaction Datasets} (Tr.Age, Tr.Gender). They are aimed at predicting age and gender \cite{ageData2020, genderData2020}. For each transaction we have the Merchant Category Codes and the decile of transaction amounts. 
Sequences of transactions provide inputs for predicting a target.

\begin{table}[]
\centering
\begin{tabular}{lcccc}
\toprule
\multicolumn{1}{l}{NLP datasets} & \textbf{AG}   & \textbf{TREC}   & \textbf{SST-2}     & \textbf{MR} \\ \midrule
FGSM              & \underline{0.66} / 0.26   & \underline{0.62} / 0.02     &  \underline{0.63} / 0.1        & \underline{0.57} / 0.03 \\
DeepFool              & 0.48 / 0.24   & 0.42 / 0.01     &  0.59 / 0.08        & 0.52 / 0.03 \\
HotFlip              & \textbf{0.78} / 0.39   & \textbf{0.75} / 0.27     & \textbf{0.84} / 0.14        & \textbf{0.62} / 0.21 \\
SamplingFool (ours)     & 0.49 / \underline{0.47}   & 0.40 / 0.37     & 0.44 / 0.41        & 0.37 / \underline{0.34} \\
DILMA, $\beta = 0$   (ours)     & 0.38 / 0.32   & 0.52 / \underline{0.52}     & 0.51 / 0.39        & 0.36 / \underline{0.34} \\
DILMA   (ours)           & 0.45 / 0.40   & 0.50 / 0.44     & 0.43 / \underline{0.43}        & 0.38 / \underline{0.34} \\
DILMA w/ sampling  (ours) & 0.59 / \textbf{0.54}   & 0.60 / \textbf{0.56}     & 0.52 / \textbf{0.51}        & 0.47 / \textbf{0.40} \\ \toprule
\multicolumn{1}{l}{Other datasets} & \textbf{EHR} & \textbf{Tr.Age} & \textbf{Tr.Gender} &             \\ \midrule
FGSM              & 0.17 / 0.04   & 0.18 / 0.17     &  \underline{0.92} / 0.59        \\
DeepFool              & 0.19 / 0.04   & 0.42 / 0.4     &  0.67 / 0.26  \\
HotFlip              & \textbf{0.21} / 0.03      & \underline{0.83} / 0.68        & \textbf{0.97} / 0.10           &             \\
SamplingFool  (ours)    & 0.01 / 0.01      & \textbf{0.85} / \textbf{0.84}        & 0.72 / \underline{0.71}           &             \\
DILMA, $\beta = 0$    (ours)           & 0.05 / \underline{0.05}      & 0.57 / 0.56        & 0.45 / 0.46           &             \\
DILMA   (ours)           & 0.06 / \underline{0.05}      & 0.60 / 0.59        & 0.46 / 0.46           &             \\
DILMA w/ sampling  (ours) & 0.06 / \textbf{0.06}      & 0.76 / \underline{0.75}        & 0.83 / \textbf{0.82}          &             \\ \hline
\end{tabular}
\caption{NAD metric ($\uparrow$) before/after adversarial training on 5000 examples. The best values are in \textbf{bold}, the second best values are \underline{underscored}. 
DILMA is resistant to adversarial training.}
\label{table:main_results}
\end{table}

\subsection{Metrics}

To create an adversarial attack, changes must be applied to the initial sequence. A change can be done either by inserting, deleting, or replacing a token in some position in the original sequence. In the $WER$ calculation, any change to the sequence made by insertion, deletion, or replacement is treated as $1$. Therefore, we consider the adversarial sequence to be perfect if $WER = 1$ and the target classifier output has changed. For the classification task, Normalised Accuracy Drop (NAD) is calculated in the following way:
\[
    NAD(A) = \frac{1}{N} \sum_{i=1}^N \frac{\mathbf{1}\{ C^t(\vecX_i) \neq C^t(\vecX'_i)) \}}{WER(\vecX_i, \vecX'_i)},
\]
where $\vecX' = A(\vecX)$ is the output of an adversarial generation algorithm for the input sequence $\vecX$, $C^t(\vecX)$ is the label assigned by the target classification model, and $WER(\vecX', \vecX)$ is the Word Error Rate. The highest value of NAD is achieved when $WER(\vecX_i', \vecX_i) = 1$ and $C(\vecX_i) \neq C(\vecX'_i)$ for all $i$. Here we assume that adversaries produce distinct sequences and $WER(\vecX_i, \vecX'_i) \geq 1$.

\subsection{Adversarial attack quality}

Results for the considered methods are presented in Table~\ref{table:main_results}.
We demonstrate not only the quality of attacks on the initial target classifier, but also the quality of attacks on the target classifier after its re-training with additional adversarial samples added to the training set.
After re-training the initial target classifier, HotFlip cannot provide reasonable results, whilst our methods perform only slightly worse than before re-training and significantly better than HotFlip and other approaches.
In case of the \textbf{Tr.Age} dataset, SamplingFool works better than DILMA because of the overall low quality of the target classifier.
We provide additional metrics in supplementary materials.

\subsection{Discriminator defense}

We have also considered another defense strategy --- discriminator training~\cite{xu2019adversarial}. We have trained an additional discriminator on $10,000$ samples of the original sequences and adversarial examples. The discriminator should detect whether an example is normal or adversarial.
The discriminator has a GRU architecture and has been trained for $5$ epochs using the negative log-likelihood loss.

A high ROC AUC of the discriminator means easy detection of adversarial attacks.
The results for the considered methods are presented in Table~\ref{table:detector_metrics}.
The discriminator ROC AUC for SamplingFool and DILMA is only slightly better than the ROC AUC of a random classifier $0.5$.
SamplingFool uses original language model without any parameter tuning, so it is even harder to detect.

\begin{table}[]
\centering
\begin{tabular}{@{}lcccc@{}}
\toprule
\multicolumn{1}{l}{NLP datasets} & \textbf{AG}  & \textbf{TREC}   & \textbf{SST-2}     & \textbf{MR} \\ \midrule
FGSM              & 0.96         & 0.71            & 0.98               & 0.94        \\
DeepFool              & 0.89         & 0.71            & 0.97              & 0.94        \\
HotFlip              & 0.99         & 0.96            & 0.99               & 0.98        \\
SamplingFool  (ours)    & \textbf{0.59}         & \textbf{0.60}            & \textbf{0.60}               & \textbf{0.60}        \\
DILMA, $\beta = 0$    (ours)          & 0.71         & 0.80            & 0.68               & 0.70        \\
DILMA    (ours)          & 0.70         & 0.71            & 0.70               & \underline{0.68}        \\
DILMA w/ sampling  (ours) & \underline{0.66}         & \underline{0.69}            & \underline{0.67}               & \textbf{0.60}        \\ \midrule
\multicolumn{1}{l}{Other datasets} & \textbf{EHR} & \textbf{Tr.Age} & \textbf{Tr.Gender} & \textbf{}   \\ \midrule
FGSM              & 0.96         & \underline{0.61}            & 0.98               &             \\
DeepFool              & \underline{0.95}         & 0.87            & 0.96               &             \\
HotFlip              & 0.99         & 0.99            & 0.99               &             \\
SamplingFool   (ours)   & \textbf{0.52}         & 0.69            & \underline{0.69}               &             \\
DILMA, $\beta = 0$   (ours)           & 0.98         & 0.71            & \textbf{0.65}              &             \\
DILMA   (ours)           & 0.98         & 0.63            & 0.83               &             \\
DILMA w/ sampling  (ours) & 0.97         & \textbf{0.57}            & 0.83               &             \\ \bottomrule
\end{tabular}
    \caption{ROC AUC scores ($\downarrow$) for Adversarial Discriminator as a countermeasure against adversarial attacks: a binary classification ``adversary vs. non-adversary''. The best values are in \textbf{bold}, the second best values are \underline{underscored}.}
    \label{table:detector_metrics}
\end{table}

\section{Conclusion}

Constructing adversarial attacks for categorical sequences is a challenging problem. Our idea is to combine sampling from a masked language model (MLM) with tuning of its parameters to produce truly adversarial examples. To tune parameters of the MLM we use a loss function based on two differentiable surrogates --- for a distance between sequences and for a classifier. This results in the proposed DILMA approach. If we only use sampling from the MLM, we obtain a simple baseline SamplingFool.

To estimate the efficiency of adversarial attacks on categorical sequences we have proposed a metric combining the WER and the accuracy of the target classifier. For considered applications that include diverse datasets on bank transactions, electronic health records, and NLP datasets, our approaches show a good performance. 
Moreover, in contrast to competing methods, our approaches win over common strategies used to defend from adversarial attacks.

\section{Broader impact}

One of the biggest problems with Deep Learning (DL) models is that they lack robustness. An excellent example is provided by adversarial attacks: a small perturbation of a target image fools a DL classifier, which predicts a wrong label for the perturbed image. The original gradient-based approach to adversarial attacks is not suitable for categorical sequences (e.g. NLP data), as the gradients are not easy to obtain. Existing approaches to adversarial attacks for categorical sequences often fail to keep the meaning and semantics of the original sequence. 

Our approach is better at this task, as it leverages modern language models like BERT. With the help of our approach, one can generate meaningful adversarial sequences that are persistent against adversarial training and defense strategies based on detection of adversarial examples. This approach poses an important question to society: can we delegate the processing of sequential data to AI? An example of a malicious use of such an approach would be an attack on a model that detects Fake news in major social networks, as an adversarial change undetectable to a human eye can render that model useless.

We hope that our work will have a broad impact, as we have made our code and all details of our experiments available to ML community. Moreover, it is easy to use, because it requires only a masked language model to work.

\section{Acknowledgements}

We thank the Skoltech CDISE HPC Zhores cluster staff for computing cluster provision.

\appendix

\section{Appendix overview}
\label{sec:appendix_intro}

In these supplementary notes we provide information and results additional to the main text of the article "Differentiable language model adversarial attacks on categorical sequence models".
The supplementary notes consist of two sections.
\begin{itemize}
    \item More detailed descriptions of models used and their training process are given in Section~\ref{sec:models}.
    \item Descriptions of additional experiments are given in Section~\ref{sec:experiments}.
\end{itemize}

\section{Models}
\label{sec:models}

In this section we provide a more detailed description of the considered models and their training, so one can reproduce the results in the article and reuse the introduced models.
We start with the description of classifiers used in Subsection~\ref{sec:classifiers_sup}, then we describe our masked language model in Subsection~\ref{sec:mlm} and the Deep Levenstein model in Subsection~\ref{sec:deep_lev}.

As we aim at adversarial attacks on existing models, we select hyperparameters using values and settings from approaches described in the literature where these models perform well.
Metrics obtained in performed experiments suggest that it is a reasonable choice and other hyperparameter choices cannot significantly improve the quality of considered models.
For our attack method we select hyperparameters using grid search.
A more detailed description of the used procedure for the selection of hyperparameters is given in Section~\ref{sec:hyperparameters}.

\subsection{Sequence classifiers}
\label{sec:classifiers_sup}

To train all models in this work we use  Adam optimiser \cite{kingma2014adam} with learning rate $0.001$ and batch \linebreak size $64$. 

\paragraph{Architecture}

In all experiments we use two classifiers: a target classifier $C^t(\vecX)$ that we attack and a substitute classifier $C(\vecX)$ that provides differentiable surrogate classifier scores. 

\textbf{The target classifier} is a Gated Recurrent Unit (GRU) RNN classifier~\cite{chung2014empirical}. We apply it in a combination with a learned embedding matrix layer $E$ of shape $d \times e$, where $d$ is the dictionary size and $e$ is the embedding dimension. We use $e = 100$, hidden state size of GRU $h = 128$, number of GRU layers $l = 1$, and dropout rate $r = 0.1$. We use a bi-directional version of GRU, averaging the GRU's output along the hidden state dimension and passing the resulting vector through one fully-connected layer.
As we show in the main paper, these settings provide a reasonable performance for considered problems.

\textbf{The substitute classifier} is a Convolutional Neural Network (CNN) for Sentence Classification proposed in~\cite{kim2014convolutional}. In our experiments, the CNN consists of multiple convolution layers and max pooling layers. The CNN has one convolution layer for each of the n-gram filter sizes. Each convolution operation gives out a vector of size $num\_filters$. We use $[3, 5]$ n-gram filter sizes with $num\_filters = 8$. The size of the learned embedding matrix $E$ remains the same $e = 100$.
The dropout ratio during training is set to $0.1$.
For this classifier the accuracy values provided in the main paper are also high.

\paragraph{Training}
As we consider various multiclass classification problems, we train both models using the cross-entropy loss for $50$ epochs.
If no improvement on the validation set occurs during $3$ epochs, we stop our training earlier and use the model, which is the most accurate on the validation set. So for most of datasets we run around $8$ epochs.

The substitute model has access only to $50\%$ of the data, whilst the target model uses the whole dataset. We split the dataset into two parts with stratification. Results are shown in Subsection \ref{sec:experiments}. Both models demonstrate comparable results for all datasets.

\begin{figure}[t]
\centering \includegraphics[width=0.5\textwidth]{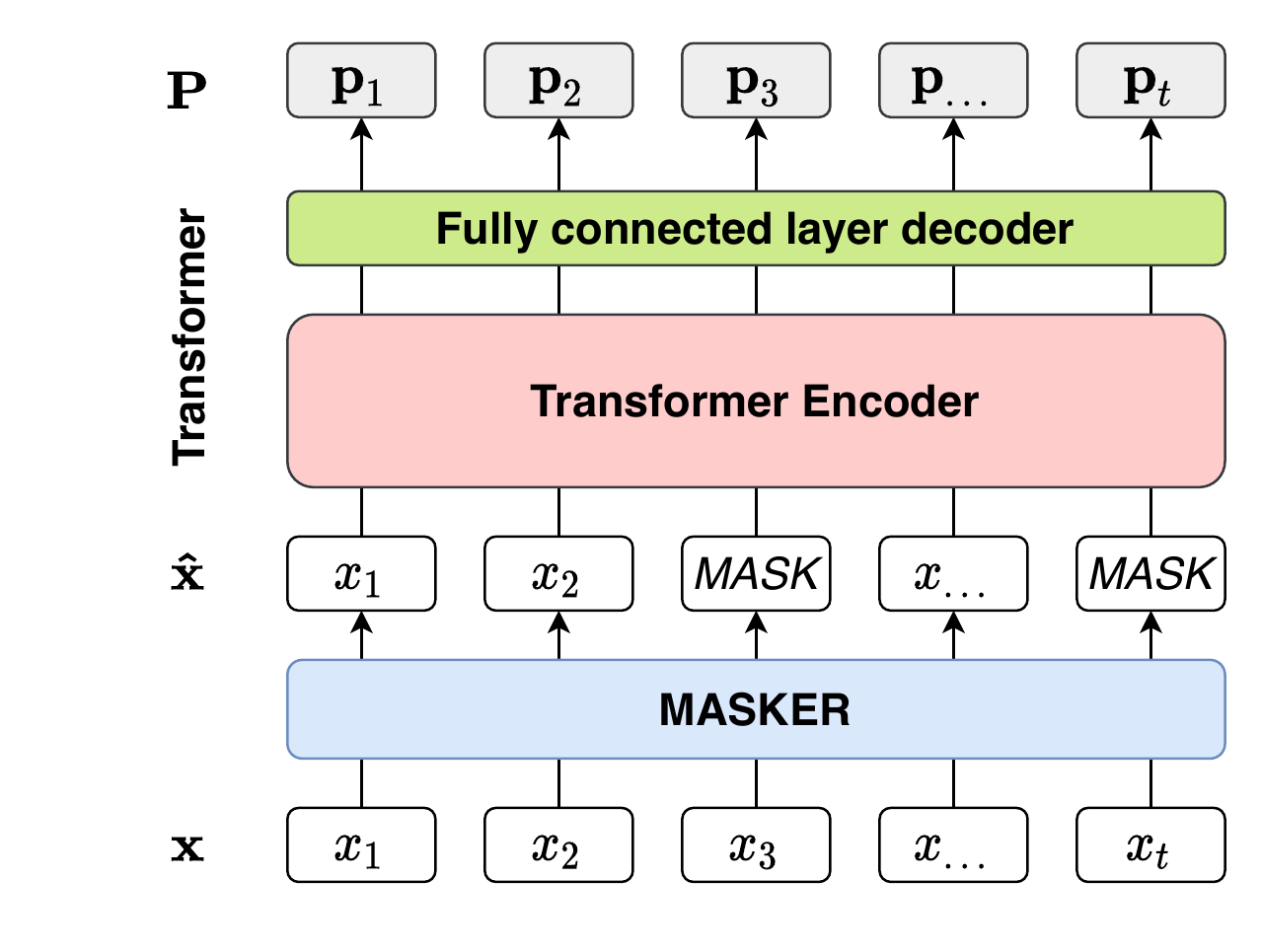}
\caption{Masked Language Model architecture}
\label{fig:mlm_architecture}
\end{figure}

\subsection{Transformer Masked Language Model}
\label{sec:mlm}

\paragraph{Masked Language Model architecture}
We use a 4-layer~\cite{vaswani2017attention} transformer encoder Masked Language Model (MLM) with the embedding of dimension of $64$ and $4$ attention heads. 

To train the Masked Language Model, we randomly replace or mask some of the tokens from the input, and the objective is to predict the original dictionary id of the masked sequence. The MLM can be seen as a generative model that captures the distribution of the data and can estimate probabilities $p(\vecX' | \vecX)$ and $p(\vecX)$~\cite{mlm_unsupervised}. 

In our experiments, we follow the ideas from \cite{devlin2018bert} and train a BERT-like LM model. The architecture of the model is shown in \autoref{fig:mlm_architecture}.
\begin{enumerate}
    \item The initial input to our model is a sequence $\vecX = (x_1, x_2, \ldots, x_t)$.
    \item The masking layer randomly adds noise to the sequence by masking and replacing some tokens $\vecX \rightarrow \hat{\vecX}$.
    \item We pass $\hat{\vecX}$ to the multi-layer bidirectional Transformer encoder \cite{vaswani2017attention} and receive a hidden representation $\vecH_i$ for each token in the sequence.
    \item Then we pass the resulting sequence of hidden representations $\vecH_i$, $i=1, \ldots, t$ through one fully-connected layer, where the output shape equals the dictionary size $d$, to get logits $\matP = \{\vecP_1, \ldots, \vecP_{t}\} \in \mathbb{R}^{d \cdot t}$ for each index $1, \ldots, t$, where $d$ is the size of the dictionary of tokens.
    \item On top of each logit vector $\vecP_i$ we can apply argmax to get for each token $i=1, \ldots, t$ the index  $x'_i$ with the maximum logit value. 
    \item Using logits $P$ we can estimate $p(\vecX' | \vecX)$.
\end{enumerate}

\paragraph{Training} 
We train the transformer MLM in the way similar to \cite{devlin2018bert}.
For a sequence $\vecX$, BERT first constructs a corrupted version $\hat{\vecX}$ by randomly replacing tokens with a special symbol [MASK] or other tokens.
Then the training objective is to reconstruct masked tokens $\bar{\vecX}$ from $\hat{\vecX}$.

The masking module changes tokens to [MASK] with a $50\%$ probability and replaces tokens with other random tokens with a $10\%$ probability to get $\hat{\vecX}$ from $\vecX$.
The output probabilities for a token at $i$-th place are:
\[
        q_{ij} = \frac{\exp \left(p_{ij} \right)}{\sum_{k = 1}^d \exp \left(p_{ik} \right)}.
\]
As the loss function we use the cross-entropy loss for generated tokens, or in other words we maximise $p_{\vecT}(\vecX | \hat{\vecX})\to \max_{\vecT}$ with respect to the parameters of the Transformer $\vecT$.
We obtain a single MLM for NLP datasets by aggregating all  available data and training the MLM for $50$ epochs. For non-NLP datasets, we train separate MLMs for the same number of epochs.

\begin{figure}[t]
\centering \includegraphics[width=0.7\textwidth]{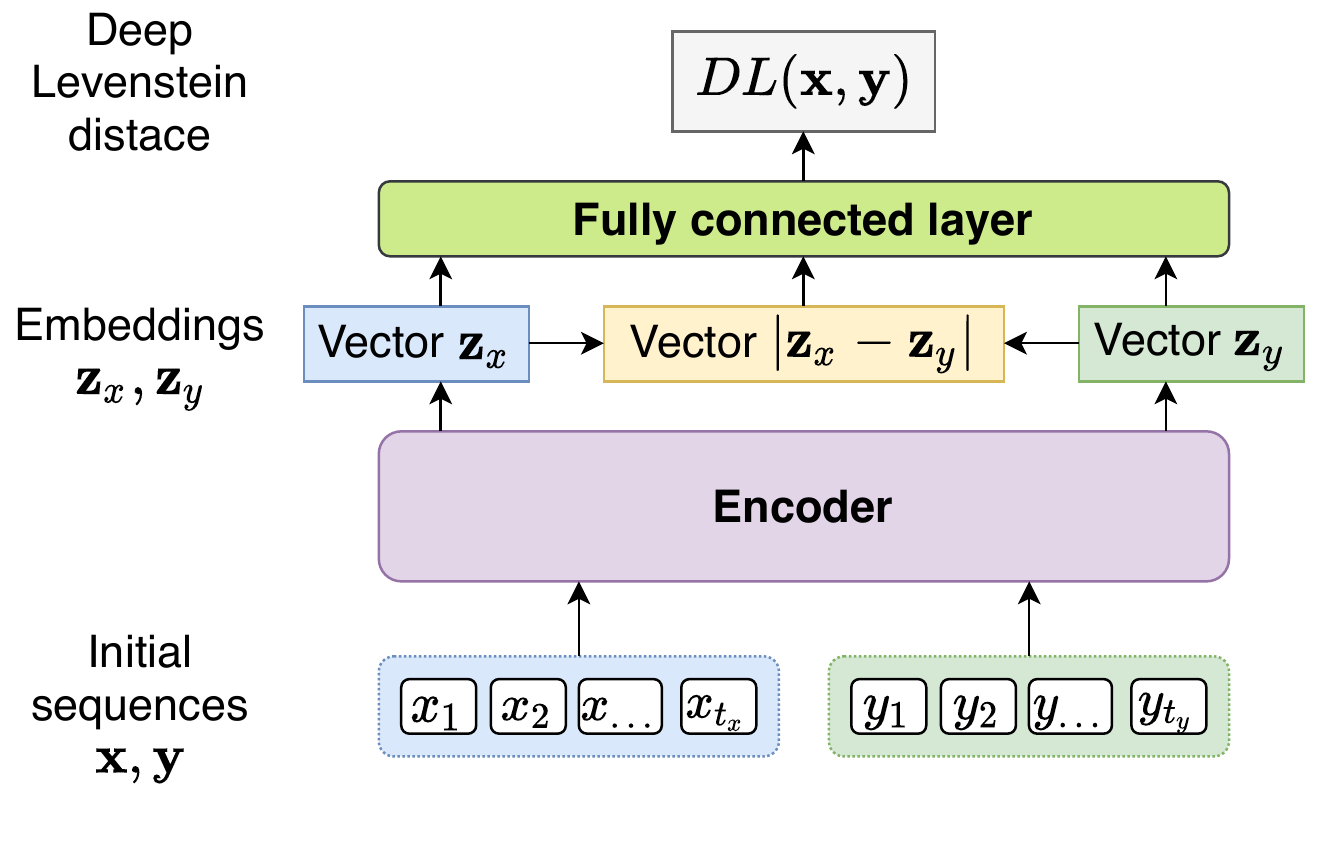}
\caption{Deep Levenstein Model architecture}
\label{fig:deep_lev_architecture}
\end{figure}

\subsection{Deep Levenstein}
\label{sec:deep_lev}

\paragraph{Architecture}
The Deep Levenstein model $DL(\vecX, \vecY)$ receives two sequences $(\vecX, \vecY)$. It encodes them into a dense representation of fixed length $l$ using the shared encoder $\vecZ_\vecX = E(\vecX)$, $\vecZ_\vecY = E(\vecY)$. Then it concatenates the representations and the absolute difference between them in a vector $(\vecZ_\vecY, \vecZ_\vecY, |\vecZ_\vecY - \vecZ_\vecY|)$ of length $3l$. 
At the end, the model uses a fully-connected layer.
The architecture is similar to the one proposed in~\cite{dai2020convolutional}. 
The schematic description of the architecture is presented in Figure~\ref{fig:deep_lev_architecture}.

\paragraph{Sample generation and training}
To estimate the parameters of the encoder and the fully connected layer we use the $L_2$ loss between true and model values of the WER distance.
We form a training sample of size of about two million data points by sampling sequences and their modifications from the training data.
To collect the training data for constructing $DL(\vecX, \vecX')$, we use sequences randomly selected from a training dataset.
As a result, we get pairs, where each pair contains a sequence and a close but different sequence obtained after application of the masking.
We also add pairs with dissimilar sequences, where each sequence is randomly selected from the training data to get a better coverage. It total we generate, around two million of pairs.

\subsection{Hyperparameters selection}
\label{sec:hyperparameters}

We use the following procedure to select hyperparameters for the DILMA attack.
\begin{enumerate}
    \item Try each combination of hyperparameters from the grid defined as a Cartesian product for values in Table~\ref{table:hyperparameters}
    \item Select the vector of hyperparameters with the best $NAD$ metric.
\end{enumerate}

\begin{table}[h!]
    \centering
    \begin{tabular}{ll}
    \hline
    Hyperparameter & Considered values \\
    \hline
    Learning rate $\alpha$ & $0.01, 0.05, 0.1, 0.5, 1$ \\
    Coefficient $\beta$ in loss function $L$ & $0.5, 1, 3, 5, 10$ \\
    Number of Gumbel samples $m$ & $1, 2, 3, \ldots, 10$ \\
    Sampling temperature $\tau$ & $1, 1.1, \ldots, 2$ \\
    Subset of updated parameters $\vecT$ & all (0), only linear (1), linear and 3rd layer (2), \\
                                           & linear and 3rd and 2nd layer (3) \\
    \hline     
    \end{tabular}
    \caption{Set of possible hyperparameters during grid search}
    \label{table:hyperparameters}
\end{table}

For the considered datasets we obtain the hyperparameters presented in Table~\ref{table:selected_hyperparameters}.
For DILMA with sampling we use the same hyperparameters.
An ablation study for the considered hyperparameters is presented in Section~\ref{sec:hyperparameters_experiment}.

\begin{table}[h!]
    \centering
    \begin{tabular}{lccccc}
    \hline
    Dataset & $\alpha$ & $\beta$ & $m$ & $\tau$ & Subset ind. \\
    \hline
    AG        & $0.01$ & $1$  & $1$  & $1.5$ & 2 \\
    MR        & $0.01$ & $10$ & $3$  & $1.8$ & 1 \\
    SST       & $0.01$ & $5$  & $1$  & $1.5$ & 2 \\
    TREC      & $0.01$ & $5$  & $10$ & $1.8$ & 1 \\
    EHR       & $0.1$  & $5$  & $3$  & $1.8$ & 1 \\
    Tr.Age    & $1$    & $1$  & $10$ & $1.2$ & 2 \\
    Tr.Gender & $0.1$  & $10$ & $10$ & $2.0$ & 2 \\
    \hline     
    \end{tabular}
    \caption{Selected hyperparameters for each dataset}
    \label{table:selected_hyperparameters}
\end{table}

\subsection{Hyperparameters for other attacks}
\label{sec:hyp_other}

For HotFlip there are no hyperparameters.
FSGM and DeepFool also have a small number of hyperparameters. The number of steps is set to $10$ for both of them and the coefficients before the step direction are set to $0.1$ for FGSM and $1.05$ for DeepFool.
We have tried a number of values, and have selected the best ones according to NAD, whilst the quality difference has been almost negligible. 

\begin{table}[]
\centering
\begin{tabular}{lcccc}
\toprule
\multicolumn{1}{l}{NLP datasets} & \textbf{AG}   & \textbf{TREC}   & \textbf{SST-2}     & \textbf{MR} \\ \midrule
FGSM                      & \textbf{1.01}   & \textbf{0.98} & 1.24 & \textbf{1.00}  \\
DeepFool                  & \underline{0.97}   & \underline{0.92} & \underline{1.18} & \underline{0.98} \\
HotFlip                   & 1.26   & 1.29 & \textbf{0.96}  & 1.04 \\
SamplingFool (ours)       & 1.37    & 1.64  & 1.24 & 1.96  \\
DILMA   (ours)            & 2.10    & 1.53  & 1.91 & 2.62  \\
DILMA w/ sampling  (ours) & 1.74    & 1.32  & 1.54 & 1.97  \\
\toprule
\multicolumn{1}{l}{Other datasets} & \textbf{EHR} & \textbf{Tr.Age} & \textbf{Tr.Gender} &             \\ \midrule
FGSM                      & \underline{1.42}   & 0.33 & \textbf{1.00}  \\
DeepFool                  & \underline{1.42}   & \underline{0.70} & 0.86   \\
HotFlip                   & 1.91   & 1.67 & \underline{1.01}   \\
SamplingFool  (ours)      & \textbf{1.28}   & \textbf{1.12} & 1.33   \\
DILMA   (ours)            & 3.98   & 2.28 & 2.70   \\
DILMA w/ sampling  (ours) & 4.06   & 1.51 & 1.91   \\
\hline
\end{tabular}
\caption{The average WER metric (the closer to $1$, the better) for FGSM, DeepFool, HotFlip, SamplingFool, DILMA and DILMA w/ sampling adversarial algorithms. As WER is the same before and after retraining of the main classifier, we provide only one number. The best values are in \textbf{bold}, the second best values are \underline{underscored}. The best value of WER is one: we change one token, and get an adversarial example, so we consider absolute difference between  $1$ and observed WER for highlighting  best approaches.}
\label{table:wer_results}
\end{table}

\begin{table}[]
\centering
\begin{tabular}{lcccc}
\toprule
\multicolumn{1}{l}{NLP datasets} & \textbf{AG}   & \textbf{TREC}   & \textbf{SST-2}     & \textbf{MR} \\ \midrule
FGSM                      & 0.46 / 0.43   & 0.32 / 0.02 & \underline{0.64} / 0.02	 & 0.23 / 0.01 \\
DeepFool                  & 0.37 / 0.17   & 0.27 / 0.00 & 0.58 / 0.05 & 0.20 / 0.01 \\
HotFlip                   & \textbf{0.62} / 0.26   & \underline{0.53} / 0.17 & \textbf{0.69} / -0.05 & \underline{0.27} / -0.04 \\
SamplingFool (ours)       & 0.34 / 0.31   & 0.28 / 0.25 & 0.48 / 0.30 & 0.16 / 0.14 \\
DILMA   (ours)            & 0.48 / \underline{0.47}   & 0.52 / \underline{0.36} & 0.60 / \textbf{0.46} & \textbf{0.28} / \underline{0.19} \\
DILMA w/ sampling  (ours) & \underline{0.52} / \textbf{0.49}   & \textbf{0.56} / \textbf{0.41} & 0.63 / \underline{0.43} & \underline{0.27} / \textbf{0.20} \\
\toprule
\multicolumn{1}{l}{Other datasets} & \textbf{EHR} & \textbf{Tr.Age} & \textbf{Tr.Gender} &             \\ \midrule
FGSM                      & 0.34 / 0.22   & 0.02 / 0.01 & \underline{0.42} / 0.14  \\
DeepFool                  & \underline{0.38} / 0.20   & 0.07 / 0.05 & 0.30 / 0.01  \\
HotFlip                   & \textbf{0.61} / 0.23   & \textbf{0.20} / -0.17 & \textbf{0.48} / 0.16  \\
SamplingFool  (ours)      & 0.04 / 0.03   & 0.15 / \underline{0.15} & 0.27 / 0.23  \\
DILMA   (ours)            & 0.28 / \underline{0.34}   & \underline{0.16} / \underline{0.15} & 0.25 / \underline{0.27}  \\
DILMA w/ sampling  (ours) & 0.35 / \textbf{0.39}   & \underline{0.16} / \textbf{0.16} & 0.29 / \textbf{0.30}  \\
\hline
\end{tabular}
\caption{The mean probability difference metric ($\uparrow$) before/after adversarial training on 5000 examples. The best values are in \textbf{bold}, the second best values are \underline{underscored}. 
DILMA is resistant to adversarial training.}
\label{table:diff_results}
\end{table}

\section{Experiments}
\label{sec:additional_experiments}

\begin{figure}[t]
\centering \includegraphics[width=0.7\textwidth]{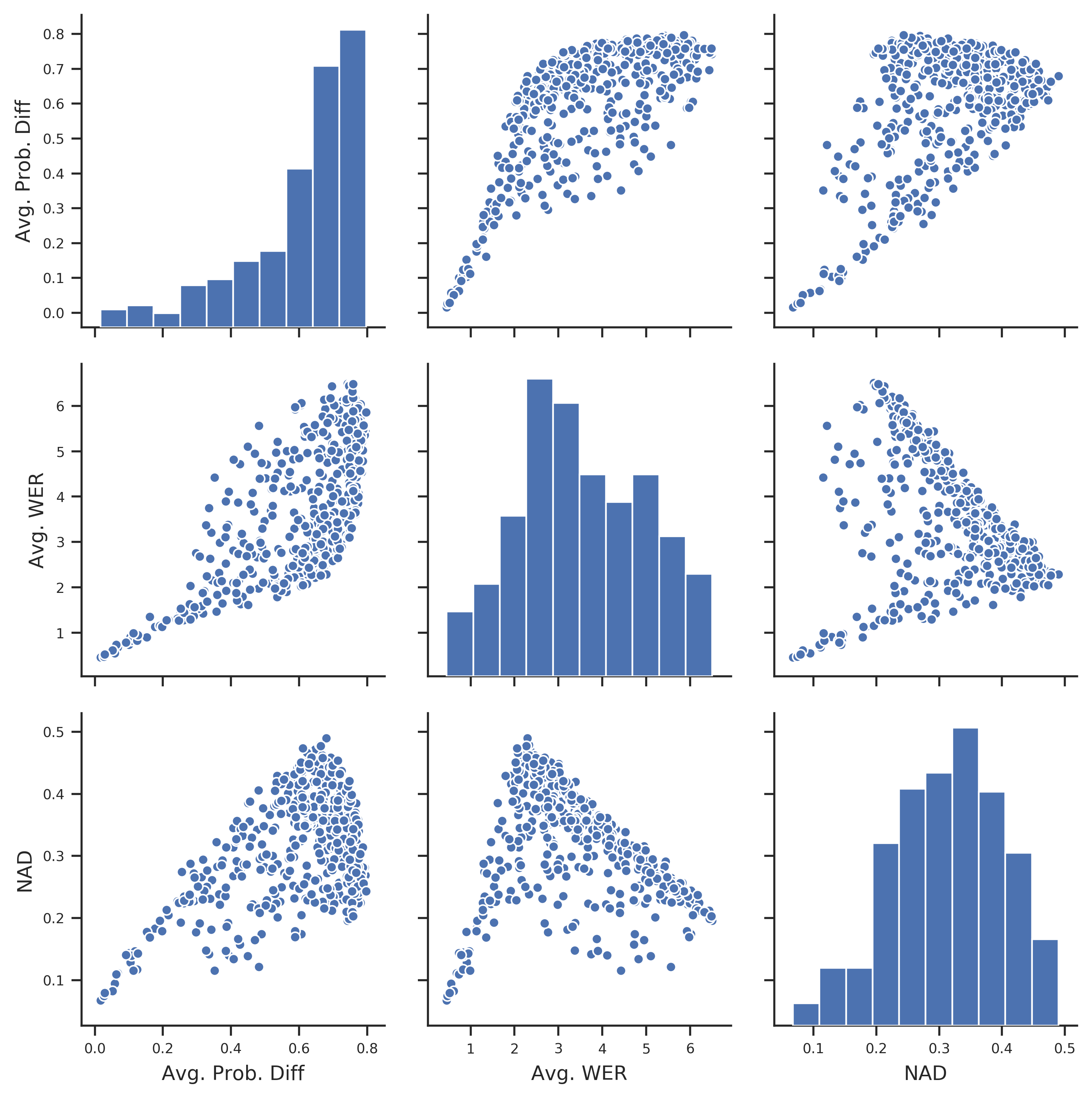}
    \caption{Hyperparameters grid search for the Tr.Gender dataset. Each point corresponds to a configuration of hyperparameters. The whole figure shows the correspondence between NAD, average WER, and average adversarial probability drop metrics.}
    \label{fig:pair_1}
\end{figure}

\begin{figure}[t]
\centering \includegraphics[width=0.7\textwidth]{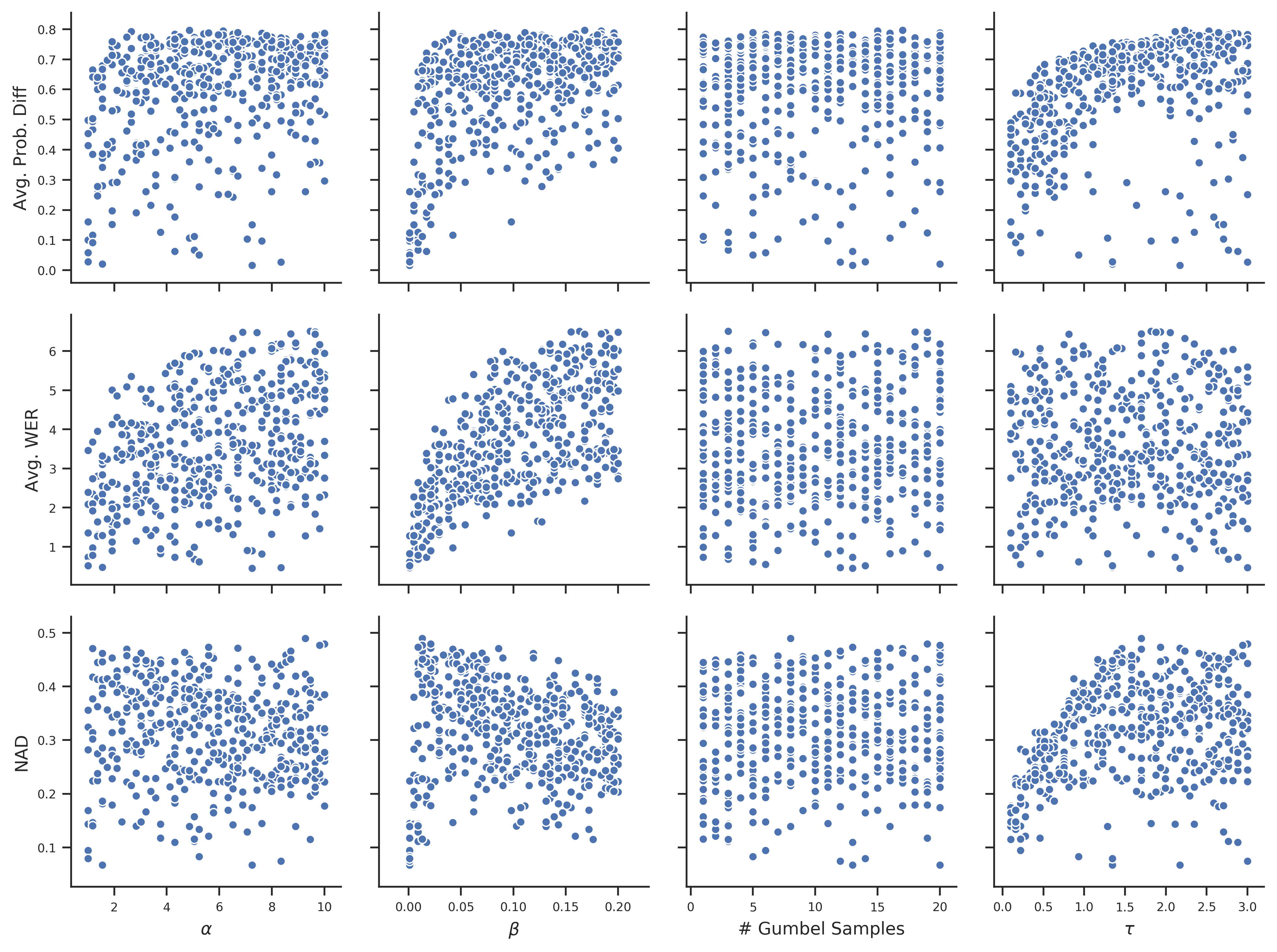}
    \caption{Hyperparameters grid search for the Tr.Gender dataset. Each point corresponds to a configuration of hyperparameters. The whole figure shows the correspondence between NAD, average WER, and average adversarial probability drop metrics for different values of hyperparameters.}
    \label{fig:pair_2}
\end{figure}

\subsection{Mean probability difference and WER metrics}

The NAD metric unifies the WER (word error rate) and true class probability difference for a particular example.
But these two metrics can also provide additional insight on the quality of considered approaches.
The WER for each pair of an initial sequence $\vecX$ and an adversarial sequence $\vecX'$ is the word error rate between them, $\mathrm{WER}(\vecX, \vecX')$.
The value of the WER should be as close as possible to 1.
For the probability difference we consider the probability $p_y(\vecX)$ of the true class $y$ for the initial sequence $\vecX$  and the adversarial sequence $p_y(\vecX')$.
The probability difference $p_y(\vecX) - p_y(\vecX')$ should be as high as possible, as the adversarial sequence should have a very low score for the true class of the initial sequence.
These two metrics are conflicting, so we typically can increase the mean probability difference by simultaneously worsening (increasing) the WER.
For DILMA, we can adjust the target WER by modifying hyperparameters (see experiments in Subsection~\ref{sec:hyperparameters_experiment}), for other approaches it is much harder to select an appropriate trade-off.

In a way similar to the main text, we measure the quality by averaging the metrics over $10,000$ adversarial examples.
In Table~\ref{table:wer_results} we provide the WER for considered approaches before and after retraining. 
In Table~\ref{table:diff_results} we provide the mean probability difference for the target class for considered approaches before and after retraining.
We see that we can make similar conclusions when comparing the quality of considered approaches.


\subsection{Ablation study}
\label{sec:hyperparameters_experiment}

In this subsection we provide an ablation study for our metric NAD and for our approach DILMA.
We vary the learning rate, the coefficient before the loss related to the classifier $\alpha$, the number of Gumbel samples, and the temperature for sampling $\tau$ to generate a grid.
Each point in Figures~\ref{fig:pair_1} and~\ref{fig:pair_2} corresponds to quality metrics for selected four  hyperparameters.
For each configuration we generated $10 000$ adversarial examples and evaluated $3$ quality metrics: the accuracy drop for a classifier, the word error rate, and our metric NAD.
We see that we can vary the WER by changing hyperparameters. 
In general for DILMA the performance depends on the choice of hyperparameters. However, we observe that we always can select the same hyperparameters setting that provides good performance across diverse datasets.

\bibliography{adversarial}

\end{document}